\documentclass[letter]{ieice}
\usepackage[pdftex]{graphicx,xcolor}
\usepackage[fleqn]{amsmath}
\usepackage{newtxtext}
\usepackage[varg]{newtxmath}

\newcommand{\AmSLaTeX}{%
 $\mathcal A$\lower.4ex\hbox{$\!\mathcal M\!$}$\mathcal S$-\LaTeX}
\def\BibTeX{{\rmfamily B\kern-.05em
 \textsc{i\kern-.025em b}\kern-.08em
  T\kern-.1667em\lower.7ex\hbox{E}\kern-.125emX}}
\makeatletter
\def\tmpcite#1{\@ifundefined{b@#1}{\textbf{?}}{\csname b@#1\endcsname}}%
\makeatother

\field{A}
\vol{98}
\no{1}
\title
      {Vision Transformer with Key-select Routing Attention for Single Image Dehazing}
\authorlist{%
 \authorentry{Lihan Tong}{n}{labelA}\MembershipNumber{}
 \authorentry{Weijia Li}{n}{labelB}\MembershipNumber{}
\authorentry{Qingxia Yang}{n}{labelA}\MembershipNumber{}
 \authorentry{Liyuan Chen}{n}{labelA}\MembershipNumber{}
 \authorentry[chenpeng@jmu.edu.cn(Corresponding author)]{Peng Chen}{n}{labelA}\MembershipNumber{}
}
\affiliate[labelA]{The authors are with the school of Ocean Information Engineering, Jimei University, Xiamen, China}
\affiliate[labelB]{The author is with the school of Computer Science, Jimei University, Xiamen, China }

\begin{document}
\maketitle

\begin{summary}
We present Ksformer, utilizing Multi-scale Key-select Routing Attention  (MKRA) for intelligent selection of key areas through multi-channel, multi-scale windows with a top-k operator, and Lightweight Frequency Processing Module (LFPM) to enhance high-frequency features, outperforming other dehazing methods in tests.
\end{summary}
\section{Introduction}\label{intro}
Single image dehazing \cite{aod,ffa-net,wu2021contrastive} aims to restore clear, high-quality images from hazy ones, essential for applications like object detection \cite{zou2023object} and semantic segmentation \cite{hao2020brief}.
Traditional methods \cite{dcp,zhu2014single,berman2016non} may not give ideal dehazing results because they can't cover all scenarios \cite{zhang2022deep}. With the rise of deep learning, convolutional neural networks (CNNs) \cite{ffa-net,wu2021contrastive,cui2022selective} have been widely applied to image dehazing and have achieved good results. However, because CNNs cannot capture long-range dependencies, this limits further improvement in dehazing effects. Recently, Transformers \cite{parmar2018image,chen2021pre,liu2021swin,ke2021musiq} have been widely used in computer vision tasks because they can capture long-range dependencies. However, they have a problem where their computational complexity is proportional to the square of the image resolution. Many efforts \cite{liu2021swin,wang2022uformer,zamir2022restormer,qiu2023mb} have been made to address this issue by introducing handcrafted sparsity. But the sparsity added by hand doesn't relate to the content, causing some loss of information. 

We propose Ksformer, which is made up of MKRA and LFPM. MKRA estimates queries in windows of different sizes and then uses a top-k operator to select the most important k queries. This approach enhances computational efficiency and incorporates content-aware capabilities. Meanwhile, multi-scale windows adeptly manage blurs of varying sizes. LFPM employs lightweight parameters to extract spectral features.
The contributions of this work are summarized as:
\begin{itemize}
    \item Ksformer is content-aware, selecting key-value pairs with important information to minimize content loss, while also capturing long-range dependencies and reducing computational complexity.

\item Ksformer extracts spectral features with ultra-lightweight parameters, performing MKRA in both spatial and frequency domains and then fusing them, which narrows the gap between clean and hazy images in terms of both space and spectrum.

 \item Ksformer achieves a PSNR of 39.4 and an SSIM of 0.998 with only 5.8M parameters, which is significantly better than other state-of-the-art methods.
\end{itemize}
\begin{figure*}[t!]
    \centering
    \includegraphics[width=0.92\linewidth]{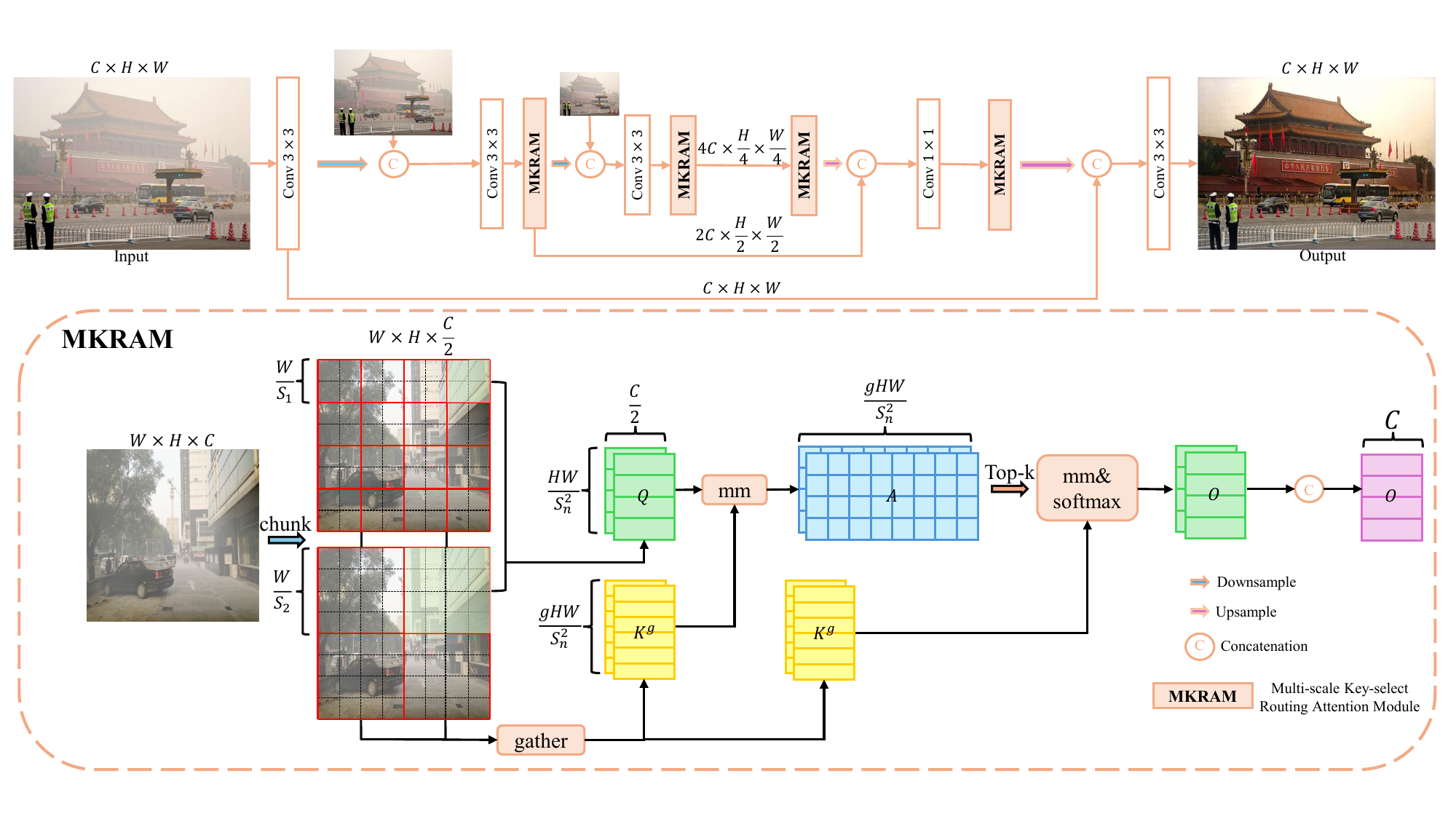}
    \caption{ The architecture of the proposed  Ksformer.}
    \label{Fig:network}
\vspace{-0.4cm}
\end{figure*}
\begin{figure}[t!]
	\centering
	\includegraphics[width=8cm]{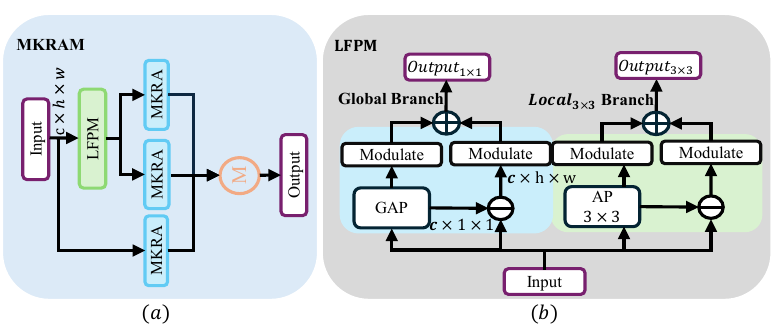}
	\caption{(a) is the architecture of the proposed  MKRAM. (b) is the architecture of the proposed LFPM. }
	\label{Fig:MFE}
	\vspace{-0.6cm}
\end{figure}
\section{Method}    
\subsection{Image Dehazing}
 We use three encoders and three decoders and downsample by $4 \times 4$ for a compact model. We use Multi-scale Key-select Routing Attention Module (MKRAM) only in the smaller dimensions to reduce computational complexity. To lower the difficulty of training \cite{mao2021deep,tu2022maxim}, we strengthen the exchange of information between layers and use skip connections at both the feature and image levels. 
\subsection{Multi-scale Key-select Routing Attention}
 MKRA uses a top-k operator to select the most important key-value pairs, balancing content awareness with lower computational complexity. For any given input feature map $X \in R^{H \times W \times C}$, first, we divide it into four parts along the channel dimension, with window sizes of $2 \times 2,4 \times 4,8 \times 8,64 \times 64$. Then, it is divided into $S \times S$ non-overlapping regions. Each region contains $\frac{H W}{S^{2}}$ feature vectors. After this step, $X$ is reshaped into $X^{r} \in R^{S^{2} \times \frac{H W}{S^{2}} \times \frac{C}{4}}$. Then, we use linear projection to weight and derive $Q, K, V \in R^{S^{2} \times \frac{H W}{S^{2}} \times \frac{C}{4}}$.
\begin{equation*}
Q=X^{r} W^{q}, K=X^{r} W^{k}, V=X^{r} W^{v}, \tag{1}
\end{equation*}
Here, $W^{q}, W^{k}, W^{v} \in R^{\frac{C}{4} \times \frac{C}{4}}$ represent the weights for $Q, K$, and $V$, respectively.
We construct an Attention module to identify the areas where important key-value pairs are located. In simple terms, we use the average values of each region to derive region-level queries and keys, $Q_{r}, K_{r} \in R^{s^{2} \times \frac{C}{4}}$. Then, we derive the region-to-region importance association matrix using the following formula.
\begin{equation*}
A_{r}^{n \times n}=\operatorname{Softmax}\left(\frac{Q_{r}\left(K_{r}\right)^{T}}{\sqrt{\frac{c}{4}}}\right), \tag{2}
\end{equation*}
Here, $A_{r}^{n \times n} \in R^{S^{2} \times \frac{C}{4}}$, represents the degree of association between two regions. $n \times n$ represents the size of the window. Next, we concatenate $A_{r}^{n \times n}$ along the channel dimension to obtain $A_{r} \in R^{S^{2} \times C}$. Next, we retain the top $\mathrm{k}$ most important queries using the top-k operator and prune the association graph to derive the index matrix.
\begin{equation*}
I_{r}=\operatorname{topk}\left(A_{r}\right), \tag{3}
\end{equation*}
Here, $I_{r} \in R^{S^{2} \times K}$. So, the i-th row of $I$ contains the k indices of the most relevant regions for the i-th region. Using the importance index matrix $I_{r}$, we can capture long-range dependencies, be content-aware, and reduce computational complexity. For each query and token in region $i$, it will focus on all key-value pairs in the union of $\mathrm{k}$ important regions indexed by $I^{(i, 1)}_{r}, I^{(i, 2)}_{r}, \ldots, I^{(i, k)}_{r}$.
We collect the key and value tensors.
\begin{equation*}
K^{g}=\operatorname{gather}\left(K, I_{r}\right), V^{g}=\operatorname{gather}\left(V_{r}\right), \tag{4}
\end{equation*}
We collect the key and value tensors. Here $K^{g}, V^{g} \in$ $R^{S^{2} \times \frac{k W H}{S^{2}} \times C}$, Finally, we focus our attention on the collected key-value pairs.
\begin{equation*}
\text { Output }=\operatorname{Attention}\left(Q, K^{g}, V^{g}\right) .\tag{5}
\end{equation*}
\section{Lightweight Frequency Processing Module}
 Past approaches often employed wavelet \cite{selesnick2005dual,yang2019wavelet,chen2021all,zou2021sdwnet} or Fourier \cite{yu2022frequency} transformations to divide image features into multiple frequency sub-bands. These approaches raised the computational load for reverse transformation and didn't boost key frequency elements. To solve this, we added a lightweight module for spectral feature extraction and modulation. It efficiently splits the spectrum into different frequencies and uses a small number of learnable parameters to emphasize the most informative ones.
 \section{Multi-scale Key-select Routing Attention Module}
 As shown in Fig.\ref{Fig:MFE}, to improve model efficiency, our MKRAM module processes the spatial domain, low frequencies, and high frequencies in parallel and then fuses the three outputs.
\begin{figure*}[t!]
    \centering
    \includegraphics[width=1\linewidth]{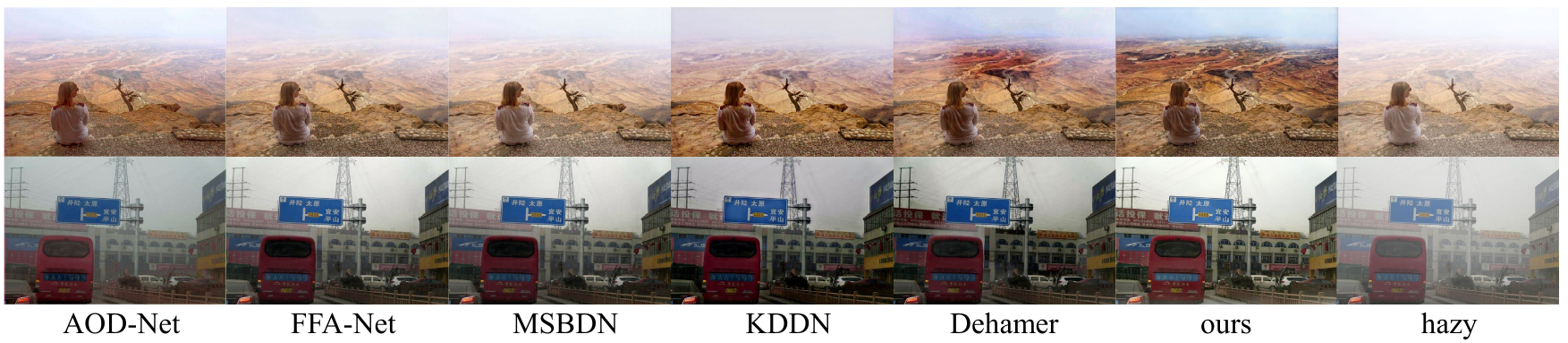}
    \caption{ Visual results comparisons on RTTS \cite{li2019benchmarking} dataset.  Zoom in for best view.}
 \label{Fig:RTTS}
\end{figure*}
\begin{figure*}[t!]
    \centering
    \includegraphics[width=1\linewidth]{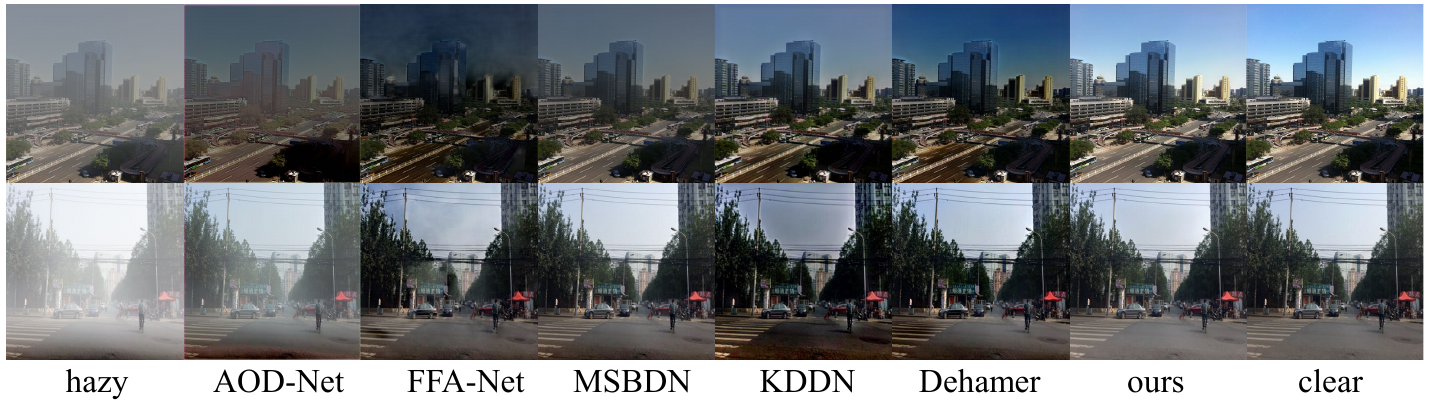}
    \caption{ Visual results comparisons on Haze4K dataset~\cite{liu2021synthetic}. Zoom in for best view.}
 \label{Fig:Haze4K}
\end{figure*}
\begin{table}[t]
	\caption{Quantitative comparisons with SOTA  methods on the RESIDE-Indoor~\cite{li2019benchmarking} and Haze4K \cite{liu2021synthetic} datasets.} 
	\label{performance_SOTS}
	\centering
 \resizebox{0.5\textwidth}{!}{
	\begin{tabular}{c|cc|cc|cc}
 \hline	
		
		 {Method}  & \multicolumn{2}{c|}{RESIDE-IN \cite{li2019benchmarking}}             & \multicolumn{2}{c|}{Haze4k \cite{liu2021synthetic}}      &  {\# Param}
  & {\# GFLOPs}
  \\ \cline{2-5}
		& \multicolumn{1}{c}{PSNR (dB)} & SSIM      & \multicolumn{1}{c}{PSNR (dB)} & SSIM      \\ \hline
		
		(ICCV'17) AOD-Net \cite{aod}                & 19.82                     & 0.818    & 17.15                     & 0.83                & 0.002M     &-          \\
		(ICCV'19) GridDehazeNet \cite{griddehazenet}          & 32.16                     & 0.984   & -                    & -       & 0.96M                &-  \\
		(AAAI'20) FFA-Net \cite{ffa-net}                & 36.39                    & 0.989   & 26.96                     & 0.95                   & 4.68M         & 288.34        \\
		(CVPR'20) MSBDN \cite{msbdn1}                  & 33.79                     & 0.984    & 22.99                 & 0.85         & 31.35M          &41.58     \\
		(CVPR'20) KDDN \cite{hong2020distilling}                   & 34.72                     & 0.985    & -                   & -        & 5.99M          &-       \\ 
		(CVPR'21) AECR-Net \cite{wu2021contrastive} & 37.17& 0.990& - & -  & 2.61M &26.10 \\ 
            (CVPR'22) Dehamer \cite{guo2022image} & 36.63& 0.988& - & - &- &59.14 \\
            (ECCV'22) PMNet \cite{ye2021perceiving} & 38.41& 0.990 &33.49 & \textbf{0.98} &18.9M &-\\
             \hline
		Ksformer(Ours)         & \textbf{39.40}                 & \textbf{0.994}
  & \textbf{33.74}            & \textbf{0.98} & 5.8M    &92.12              \\ 
  \hline	
	\end{tabular}
 }
	\vspace{-0.3cm}
\end{table}
\begin{table}[tb]%
\caption{Ablation study of our Ksformer on the Haze4k Dataset~\cite{liu2021synthetic}.}
\label{ablation_netwrok}
\begin{center}
\begin{tabular}{cc|cc}
\hline	
		Model & PSNR (dB) & SSIM \\
		\hline 
            Base(U-Net)   &25.46 &0.92\\
		Base+MKRA   &32.25 &0.94\\
		Base+LFPM  &28.52 &0.92\\
            Base+MKRA+LFPM  &33.23 &0.95\\
		Base+MKRAM (Full)  & \textbf{33.74} & \textbf{0.98} \\		
  \hline
	\end{tabular}
\halflineskip
\end{center}
\end{table}
\section{Experiments}
\subsection{Implementation Details}
During our experiments, we employ PyTorch version 1.11.0 and utilize the capabilities of four NVIDIA RTX 4090 GPUs to perform all tests. In the training phase, images are randomly cropped into 320 × 320 pixel patches. For assessing the model's computational complexity, we adopt a size of 128 × 128 pixels. The Adam optimizer is engaged for optimization, with decay rates set at 0.9 for $\beta_1$  and 0.999 for $\beta_2$. The initial learning rate is configured at 0.00015, and we apply a cosine annealing strategy for its scheduling. The batch size is maintained at 64. Through empirical determination, we set the penalty parameter $\lambda$ to 0.2 and $\gamma$ to 0.25, and we proceed with training for 80,000 iterations.

\subsection{Quantitative and Qualitative Experiments}
 Visual Comparison. To thoroughly assess our method, we tested it on both the synthetic Haze4K \cite{liu2021synthetic} dataset and the real-world RTTS \cite{li2019benchmarking} dataset. As shown in Fig.\ref{Fig:RTTS} and Fig.\ref{Fig:Haze4K}, it's clear that our method outperforms others in terms of edge sharpness, color fidelity, clarity of texture details, and handling of sky areas, whether on synthetic or real datasets. Quantitative Comparison. We quantitatively compared Ksformer with the current state-of-the-art methods on the SOTS indoor \cite{li2019benchmarking} and Haze4K \cite{liu2021synthetic} datasets. As shown in Table.\ref{performance_SOTS}, for the SOTS indoor \cite{li2019benchmarking} dataset, Ksformer achieved a PSNR of 39.40 and an SSIM of 0.994, which is a 0.09 PSNR improvement over the second-best method, and it did so with only $30\%$ of the parameter volume. For the Haze4K \cite{liu2021synthetic} dataset, Ksformer reached a PSNR of 33.74 and an SSIM of 0.98. The quantitative comparison fully demonstrates that Ksformer outperforms other state-of-the-art methods in terms of performance.
\subsection{Ablation Study} 
To prove the effectiveness of our method, we conducted an ablation study. We first built a U-Net as the base network, and then gradually added modules to the baseline. As shown in Table.\ref{ablation_netwrok}, both PSNR and SSIM improved with the step-by-step addition of modules, and the metrics reached their best values after effectively combining the modules we proposed.
\section{Conclusion}
This paper introduces Ksformer, which combines a top-k operator with multi-scale windows, giving the network the characteristics of content awareness and low complexity. At the same time, it obtains spectral features with ultra-lightweight parameters, narrowing the spectral gap between clean and foggy images. On the SOTS indoor \cite{li2019benchmarking} dataset, it achieved a PSNR of 39.4 and an SSIM of 0.994 with only 5.8M.

Although the Ksformer has a relatively small parameter count of just 5.8 million, it's unfortunate that it can't be implemented on embedded systems due to its high GFLOPs. We plan to further explore the balance between performance and computational complexity. By appropriately reducing the number of channels and modules, we aim to make the Ksformer suitable for embedded systems, allowing it to play a significant role in a broader range of fields.
\section*{Acknowledgments}
This work was supported in part by the Youth Science and Technology Innovation Program of Xiamen Ocean and Fisheries Development Special Funds (23ZHZB039QCB24),
Xiamen Ocean and Fisheries Development Special Funds (22CZB013HJ04).

\end{document}